\documentclass[11pt]{article}

\usepackage[final]{acl}

\usepackage{times}
\usepackage{latexsym}

\usepackage[T1]{fontenc}

\usepackage[utf8]{inputenc}

\usepackage{microtype}

\usepackage{graphicx}
\usepackage{subcaption}
\usepackage{booktabs} % for professional tables

\usepackage{amsmath,amssymb}
\usepackage{bm}
\usepackage{algorithm}
\usepackage{algorithmic}
\usepackage{caption}
\usepackage{wrapfig}
\usepackage{enumitem}
\usepackage{multirow,multicol}
\usepackage{dcolumn}
\usepackage[table,x11names]{xcolor}
\usepackage{xspace}

\usepackage{mathtools}
\usepackage{amsthm}

\usepackage[capitalize,noabbrev]{cleveref}

\theoremstyle{plain}

\theoremstyle{definition}

\theoremstyle{remark}

\usepackage[textsize=tiny]{todonotes}

\newcommand{\llmbench}{LLM-as-a-benchmark\xspace}
\newcommand{\llmeval}{LLM-as-an-evaluator\xspace}
\newcommand{\llmtest}{LLM-as-a-testset\xspace}

\title{When LLMs Benchmark Themselves: Deconstructing Self-Bias in Automated Evaluation}

\author{Wenda Xu\thanks{Email Correspondence: \texttt{\{wendax, swetaagrawal, freitag, dandeutsch\}@google.com}, \texttt{vzouhar@ethz.ch}}\textsuperscript{1} \quad
  Sweta Agrawal\textsuperscript{1} \quad
  Vilém Zouhar\textsuperscript{2} \quad
  Markus Freitag\textsuperscript{1} \quad
  Daniel Deutsch\textsuperscript{1} \\
  \textsuperscript{1}Google \quad
  \textsuperscript{2}ETH Zurich
}

\begin{document}
\maketitle
\vspace{-1em}
\setlength{\parskip}{0pt}
\begin{abstract}
As LLMs rapidly saturate existing benchmarks, automated benchmark creation using LLMs (\llmbench)---where a model generates test inputs (\llmtest) and evaluates outputs (\llmeval)---has gained traction as a cheap alternative to human curation. We show that this paradigm has a fundamental problem: \textit{LLM-generated benchmarks systematically favor the model that created them}. Using machine translation as our primary testbed, we find that self-bias arises from two compounding sources, \llmtest and \llmeval, and their combination amplifies the effect. Crucially, even when test data is generated with explicit diversity controls, each model's implicit stylistic tendencies produce homogeneous, model-specific outputs that inflate its own scores. Increasing source text diversity, using our proposed diversity metric, partially mitigates this bias. Self-bias is strong enough to cause each model to rank itself first, overriding the peer-consensus ordering. We confirm that the phenomenon extends to open-ended generation on the Chatbot Arena task.
\end{abstract}

\section{Introduction}
Keeping NLP benchmarks current is harder than it used to be. LLMs saturate leaderboards within months of a benchmark's release, yet building a new human-curated test set is expensive and slow---and for languages or domains where labeled data is scarce, it may simply not be feasible. Automated benchmark creation using LLMs (\llmbench)~\citep{farchi2024automaticgenerationbenchmarksreliable, maheshwari2024efficacysyntheticdatabenchmark, pombal2025zeroshotbenchmarkingframeworkflexible} has emerged as a practical response: a single model generates test inputs (\llmtest) and scores the outputs (\llmeval), producing benchmarks on demand at low cost.

The problem is that this convenience comes with a hidden cost. Prior work has shown that LLM-based evaluators tend to favor their own outputs~\citep{llm-favors-its-own-outputs, chen2025llmevaluatorspreferreason, xu-etal-2024-pride}, but the bias introduced by the test set generation step---and how it compounds with evaluator bias---has not been carefully studied. If the source texts a model generates are already tailored to its own strengths, performance scores will reflect the model's idiosyncrasies rather than its actual quality.

We study this problem and make three contributions. \textbf{(1)} We formally define self-bias using the estimator-bias framework~\citep{xu-etal-2024-pride}, decomposing it into \llmtest and \llmeval components and showing that their combination compounds the effect. Self-bias is strong enough that every model ranks itself first, overriding the peer-consensus ordering as estimated by the remaining models. \textbf{(2)} We trace self-bias to a plausible mechanistic cause: even under high-quality prompts with explicit diversity controls~\citep{pombal2025zeroshot}, each model generates repetitive, model-specific source texts---text it can then translate and score more easily. This effect is strongest in low-resource languages, where limited generation capacity causes outputs to converge into distinctive degeneration patterns. We also verify that self-bias appears in \llmbench on the Chatbot Arena task, showing the problem is not specific to translation. \textbf{(3)} We provide practical guidance for reducing self-bias: filtering source texts using our proposed diversity metric reduces bias, though the effectiveness varies across models; generating benchmark data in languages where models have high proficiency minimizes model-specific signatures; and using frontier models to benchmark open-source models remains reliable when the quality gap is large.

% \section{Preliminaries \dd{Preliminary Definitions}}
\section{Preliminaries and Self-Bias Definition}
\label{sec:definition}

\noindent\textbf{\llmtest} We automatically generate a test set using a generator model, $M_\text{test}$. For a given instruction $s$, $M_\text{test}$ produces a source text $x$ and an optional reference text $y'$. The source text $x$ is then used as a prompt for the target model under evaluation, $M_\text{target}$, which generates an output $y$. Finally, an evaluation metric computes a quality score for $y$ by comparing it against source text $x$. We investigate testset generation self-bias: a generated test set may systematically favor its generator model ($M_\text{test}$), inflating that model's scores.

\noindent\textbf{\llmeval} An evaluator model, $M_\text{evaluator}$, assesses the quality of an output $y$ generated by a target model $M_\text{target}$ in response to a prompt $x$ , assigning it a score based on predefined criteria. In our experimental setup, the prompts $x$ are sourced from human-authored benchmark datasets. We investigate evaluator self-bias: an LLM-based evaluator may assign higher scores to its own model's outputs than warranted.

\noindent\textbf{\llmbench} combines \llmtest and \llmeval: a single model generates source text $x$, prompts a target model $M_\text{target}$ to produce output $y$, and then evaluates $y$. By unifying both roles in one model, \llmbench inherits---and potentially compounds---the biases of each component.

% \section{Self-Bias Definition}

We operationalize self-bias as follows:
% Our methodology for quantifying self-bias is predicated on the statistical definition of estimator bias:
\begin{equation}
% \text{bias}_{M_i} = \underbrace{\theta_{M_i, M_i}}_\text{self-ranking} - \underbrace{\frac{1}{J} \sum_{M_o \neq M_i} \theta_{M_i, M_o}}_\text{ranking by other models}
% Vilém: just a complicated balancing
\text{bias}_{M_i} = 
\underbrace{\theta_{M_i, M_i} \vphantom{\sum_{M_o \neq M_i}}}_\text{self-ranking} 
\,\, - \,\,
\underbrace{\frac{1}{|M|-1} \sum_{M_o \neq M_i} \theta_{M_i, M_o}}_\text{ranking by other models}
\end{equation}

where $M$ is a set of models, $M_i$ is the model being evaluated, and $M_o$ are peer models.
$\theta_{M_i, M_o}$ is the \textit{ranking} of model $M_i$ as determined by model $M_o$, averaged across segments. We use rankings rather than raw scores so that self-bias is directly comparable across settings where quality metrics operate on different scales.
$\theta_{M_i, M_i}$ is the self-ranking and $\frac{1}{|M|-1} \sum_{M_o \neq M_i} \theta_{M_i, M_o}$ is the peer-consensus ranking, which we treat as a proxy for true performance. Since peer models may carry their own biases, our framework measures \emph{relative} self-preference rather than deviation from an absolute ground truth.
A negative $\text{bias}_{M_i}$ indicates that $M_i$ ranks itself more favorably than its peers do; a more negative value reflects stronger self-preference.

In our study, $M_o$ plays the role of test set generator (\llmtest), evaluator (\llmeval), or both (\llmbench).

\section{Experimental Setup}
We study self-bias through machine translation, which offers established quality metrics \citep{juraska-etal-2024-metricx, kocmi2023largelanguagemodelsstate, xu-etal-2023-instructscore, lavie-etal-2025-findings}, annually updated test sets with official human rankings for our models \citep{kocmi-etal-2025-findings} as ground truth, and the ability to study self-bias across diverse language pairs. We validate the generality of our findings on open-ended generation via the Chatbot Arena task (\Cref{sec:non_translation}). For translation, our experiments encompass six low-to-medium resource language directions: Bemba$-$English, Kurdish$-$English, Aymara$-$English, Luo$-$English, English$-$Bemba, and English$-$Aymara. We focus on low-to-medium resource languages where model quality differences are larger; on high-resource pairs, all models score near-perfectly and self-bias becomes difficult to measure (See Appendix \cref{sec:high_resource} for detailed discussion). For the Chatbot Arena task, we contrast high-resource (English, Korean) and low-resource (Bemba, Aymara) prompt languages.

\paragraph{Model selection.} We select Gemini 2.5 Pro, GPT-4.1, and Claude Opus 4 for two reasons. First, they are the top-performing systems at WMT25 \citep{kocmi-etal-2025-findings}, for which official human rankings can still distinguish their translation quality. Second, all three have been used in the zero-shot benchmarking framework of \citet{pombal2025zeroshotbenchmarkingframeworkflexible}, demonstrating their capability as LLM-as-a-benchmark generators. In preliminary experiments, multilingual open-source models (Gemma3-27B, Mistral-Large-2411, Qwen3-32B) failed to generate high-quality benchmark data for low-resource languages.

\paragraph{Evaluation setup.} We evaluate models under the three conditions defined in \Cref{sec:definition}. For \llmtest, translation quality is measured by MetricX-QE~\citep{juraska-etal-2024-metricx}, a SOTA reference-free translation metric where lower scores indicate higher quality. For \llmeval, source texts are drawn from the FLORES-200 benchmark~\citep{nllbteam2022languageleftbehindscaling} to ensure human-authored test set inputs, so that the LLM acts solely as an evaluator. For \llmbench, models generate source texts and evaluate translations; a fully reference-free generation variant is analyzed in \Cref{sec:ref_free_self_bias}. We additionally use chrF~\citep{popovic-2015-chrf} to measure textual similarity across source texts generated by different models. As described in \Cref{sec:definition}, all raw scores are normalized into per-segment system rankings to ensure comparability across settings.

For \llmtest and \llmbench, we generate 200 source texts per language direction; for \llmeval, we sample 200 FLORES examples per direction. In all cases, translations are obtained from all three LLMs. \textbf{All evaluations are reference-free.} Temperature is set to $0$ for all API calls. All prompts are provided in Appendix~\Cref{sec:prompt_setup}. Our sample size of $N{=}200$ is well justified: bootstrap resampling ($B{=}10{,}000$) shows that confidence intervals for self-rankings and peer-rankings do not overlap, confirming statistically significant self-bias at this sample size (Appendix \Cref{sec:bootstrap}). We further assess statistical significance via permutation tests ($10{,}000$ resamples) throughout; significance at $p<0.05$ is marked with $^*$ in all tables.

\section{Does the use of \llmbench introduce self-bias?}
\label{sec:self_bias_llm_as_benchmark}

We investigate self-bias using machine translation, where source and target languages can be precisely controlled. To generate our benchmark data, we employ the high-quality prompt proposed by \citet{pombal2025zeroshotbenchmarkingframeworkflexible} in their \llmbench framework. This prompt explicitly controls for diversity and quality by instructing models to generate source-reference text pairs covering diverse topics, varying in length and complexity, and targeting different styles (see Appendix~\Cref{sec:prompt_setup} for details). Despite these explicit controls, we find that self-bias persists due to implicit model-specific features in the generated texts.

\Cref{tab:self_across_languages} quantifies self-bias for Gemini-2.5-pro, GPT-4.1, and Claude-Opus-4 across XX${\rightarrow}$En directions. In our framework, a negative bias score indicates that a model ranks itself more favorably than its peers do. The consistently negative diagonal scores confirm that \textbf{these models systematically favor their own outputs} across four language directions---except Gemini on Kurdish$\to$English. We hypothesize that this exception arises because Kurdish is a relatively higher-resource language compared to Bemba, Aymara, and Luo, leading to more diverse and less model-specific source text generation (further analyzed in \Cref{sec:pitall_src_only_gen}).

\paragraph{Self-bias causes ranking flips.} To demonstrate the practical impact of self-bias, we compare the rankings produced by each model's own \llmbench. Under \llmbench, \textbf{each model ranks itself first}: as shown in \Cref{tab:ranking_flips}, the diagonal entries (bolded) are always the lowest average rankings in each column, meaning each benchmark ranks its own translations as best. The per-language bias scores in \Cref{tab:self_across_languages} confirm this pattern: the diagonal entry is consistently the most negative in each column. For example, on Bemba${\rightarrow}$English, Claude-as-benchmark ranks Claude first ($-0.617$) even though its peer models disagree. This rank-flip effect---where self-bias overrides the peer-consensus ordering---is the most direct evidence that \llmbench distorts practical model selection decisions.

\begin{table}[t]
\centering
\small
\begin{tabular}{l@{\hspace{1mm}}lm{11mm}m{11mm}m{11mm}}
% \toprule
& Bemba${\rightarrow}$En.
& \multicolumn{3}{c}{LLM-as-a-benchmark} \\
&& \textbf{Gemini} & \textbf{GPT} & \textbf{Claude} \\
\cmidrule{2-5}
\parbox[t]{2mm}{\multirow{3}{*}{\rotatebox[origin=r]{90}{Translator\hspace{-5mm}}}}
& \bf Gemini & \cellcolor{Firebrick3!58}{-}0.591$^*$ & \cellcolor{SeaGreen3!13}\phantom{-}0.161 & \cellcolor{SeaGreen3!34}\phantom{-}0.430 \\
& \bf GPT & \cellcolor{Firebrick3!22}{-}0.145 & \cellcolor{Firebrick3!26}{-}0.202$^*$ & \cellcolor{SeaGreen3!27}\phantom{-}0.347 \\
& \bf Claude & \cellcolor{SeaGreen3!8}\phantom{-}0.102 & \cellcolor{SeaGreen3!40}\phantom{-}0.515 & \cellcolor{Firebrick3!60}{-}0.617$^*$ \\
\bottomrule
\end{tabular}
\quad
\begin{tabular}{l@{\hspace{1mm}}lm{11mm}m{11mm}m{11mm}}
% \toprule
& Aymara${\rightarrow}$En.
& \multicolumn{3}{c}{LLM-as-a-benchmark} \\
&& \textbf{Gemini} & \textbf{GPT} & \textbf{Claude} \\
\cmidrule{2-5}
\parbox[t]{2mm}{\multirow{3}{*}{\rotatebox[origin=r]{90}{Translator\hspace{-5mm}}}}
& \bf Gemini & \cellcolor{Firebrick3!23}{-}0.157$^*$ & \cellcolor{SeaGreen3!14}\phantom{-}0.180 & \cellcolor{Firebrick3!12}{-}0.022 \\
& \bf GPT & \cellcolor{SeaGreen3!18}\phantom{-}0.232 & \cellcolor{Firebrick3!36}{-}0.315$^*$ & \cellcolor{SeaGreen3!7}\phantom{-}0.083 \\
& \bf Claude & \cellcolor{SeaGreen3!20}\phantom{-}0.253 & \cellcolor{SeaGreen3!29}\phantom{-}0.365 & \cellcolor{Firebrick3!60}{-}0.617$^*$ \\
\bottomrule
\end{tabular}

\bigskip

\begin{tabular}{l@{\hspace{1mm}}lm{11mm}m{11mm}m{11mm}}
% \toprule
& Luo${\rightarrow}$En.
& \multicolumn{3}{c}{LLM-as-a-benchmark} \\
&& \textbf{Gemini} & \textbf{GPT} & \textbf{Claude} \\
\cmidrule{2-5}
\parbox[t]{2mm}{\multirow{3}{*}{\rotatebox[origin=r]{90}{Translator\hspace{-5mm}}}}
& \bf Gemini & \cellcolor{Firebrick3!34}{-}0.300$^*$ & \cellcolor{SeaGreen3!10}\phantom{-}0.120 & \cellcolor{SeaGreen3!14}\phantom{-}0.180 \\
& \bf GPT & \cellcolor{SeaGreen3!30}\phantom{-}0.385 & \cellcolor{Firebrick3!51}{-}0.508$^*$ & \cellcolor{SeaGreen3!10}\phantom{-}0.123 \\
& \bf Claude & \cellcolor{SeaGreen3!15}\phantom{-}0.188 & \cellcolor{SeaGreen3!28}\phantom{-}0.352 & \cellcolor{Firebrick3!54}{-}0.540$^*$ \\
\bottomrule
\end{tabular}
\quad
\begin{tabular}{l@{\hspace{1mm}}lm{11mm}m{11mm}m{11mm}}
% \toprule
& Kurdish${\rightarrow}$En.
& \multicolumn{3}{c}{LLM-as-a-benchmark} \\
&& \textbf{Gemini} & \textbf{GPT} & \textbf{Claude} \\
\cmidrule{2-5}
\parbox[t]{2mm}{\multirow{3}{*}{\rotatebox[origin=r]{90}{Translator\hspace{-5mm}}}}
& \bf Gemini & \cellcolor{SeaGreen3!1}\phantom{-}0.005 & \cellcolor{SeaGreen3!17}\phantom{-}0.215 & \cellcolor{Firebrick3!28}{-}0.219 \\
& \bf GPT & \cellcolor{SeaGreen3!14}\phantom{-}0.173 & \cellcolor{Firebrick3!22}{-}0.150$^*$ & \cellcolor{Firebrick3!12}{-}0.023 \\
& \bf Claude & \cellcolor{Firebrick3!18}{-}0.095 & \cellcolor{SeaGreen3!40}\phantom{-}0.520 & \cellcolor{Firebrick3!44}{-}0.425$^*$ \\
\bottomrule
\end{tabular}
\caption{Self-bias (\llmbench) across four XX$\rightarrow$En directions. Diagonal entries (self-bias) are significantly more negative than peer ranking bias ($^*p<0.05$) for all models except Gemini at Kurdish$\rightarrow$En.}
\label{tab:self_across_languages}
\end{table}

\begin{table}[t]
\centering
\small
\begin{tabular}{l@{\hspace{2mm}}ccc}
\toprule
\multirow{2}{*}{\bf LLM as MT system} & \multicolumn{3}{c}{LLM-as-a-benchmark} \\
& \textbf{Gemini} & \textbf{GPT} & \textbf{Claude} \\
\midrule
\bf Gemini  & \cellcolor{SeaGreen3!30}\bf 1.222* & 1.509 & 1.457 \\
\bf GPT     & 1.525 & \cellcolor{SeaGreen3!30}\bf 1.222* & 1.506 \\
\bf Claude  & 1.564 & 1.782 & \cellcolor{SeaGreen3!30}\bf 1.123* \\
\bottomrule
\end{tabular}

\caption{Rankings converted from LLM-as-an-evaluator scores (lower$=$better), averaged over four XX$\rightarrow$En directions. Each column represents a different LLM acting as the benchmark. The diagonal entries ranks itself significantly higher than peer ranking ($^*p<0.05$).}
\label{tab:ranking_flips}
\end{table}

\paragraph{Decomposition of self-bias.} To understand what drives self-bias, we decompose \llmbench into its \llmtest component (\Cref{tab:ablation_benchmark_testset}), where models generate test sets and translation quality is scored by MetricX-QE, an independent neural metric unrelated to any of the evaluated LLMs. Permutation tests ($p<0.05$) confirm that \llmbench bias is significantly larger than \llmtest bias alone for all three models, demonstrating that testset generation bias and evaluator bias compound---the combination amplifies the distortion beyond what testset generation produces in isolation. We also verify that \llmeval, where models act as judges for translations of human-authored FLORES test data, independently exhibits self-bias (see Appendix~\Cref{sec:evaluator_bias_appendix}).

\begin{table}[t]
    \centering
    \small
    \begin{tabular}{lm{10mm}m{10mm}m{10mm}}
    \toprule
    XX${\rightarrow}$En & \bf Gemini & \bf GPT & \bf Claude \\
    \midrule
    \bf \llmtest     
    & \cellcolor{Firebrick3!18}-0.124 
    & \cellcolor{Firebrick3!35}-0.239 
    & \cellcolor{Firebrick3!13}-0.093 \\
    \bf \llmbench   
    & \cellcolor{Firebrick3!38}-0.261$^*$ 
    & \cellcolor{Firebrick3!43}-0.294$^*$ 
    & \cellcolor{Firebrick3!80}-0.550$^*$ \\
    \bottomrule
    \end{tabular}
    \caption{Self-bias under \llmtest and \llmbench (averaged XX$\rightarrow$En). \llmbench is significantly more self-biased than \llmtest for all models ($^*p<0.05$), confirming that evaluator bias compounds with testset bias. \llmeval results (using human-authored FLORES inputs) appear in Appendix~\Cref{sec:evaluator_bias_appendix}.}
    \label{tab:ablation_benchmark_testset}
\end{table}

\paragraph{Where does the self-bias stem from?} Co-generating source texts and translation references, as in \cite{pombal2025zeroshot}, can introduce two biases: (1)~a \emph{translatability bias}, where the model avoids source texts it cannot translate well, and (2)~a \emph{generation bias}, where each model generates text with its own model-specific patterns. We confirm translatability bias by comparing MetricX-QE scores for src-only vs.\ src+ref generation (\Cref{tab:ablation_translation_quality}): src+ref consistently yields better scores, indicating that \textbf{co-generation tailors sources to the model's translation strengths}.

\begin{table}[t]
    \centering
    \small
    \begin{tabular}{llm{7mm}m{7mm}m{7mm}}
    \toprule
    Lang & Setup & \bf Gemini & \bf GPT & \bf Claude \\
    \midrule
    \multirow{2}{*}{Bemba$\to$En}
    & Src-only & \cellcolor{Firebrick3!80}8.49 & \cellcolor{Firebrick3!50}5.33 & \cellcolor{Firebrick3!38}4.07 \\
    & Src+Ref  & \cellcolor{Firebrick3!26}2.73 & \cellcolor{Firebrick3!45}4.76 & \cellcolor{Firebrick3!36}3.89 \\
    \midrule
    \multirow{2}{*}{Aymara$\to$En}
    & Src-only & \cellcolor{Firebrick3!80}12.9 & \cellcolor{Firebrick3!65}10.4 & \cellcolor{Firebrick3!51}8.29 \\
    & Src+Ref  & \cellcolor{Firebrick3!47}7.54  & \cellcolor{Firebrick3!57}9.60  & \cellcolor{Firebrick3!50}8.13 \\
    \bottomrule
    \end{tabular}
    \caption{Src+Ref generation yields better translation quality than src-only (MetricX; lower is better), indicating co-generation biases sources toward the model's translation strengths.}
    \label{tab:ablation_translation_quality}
\end{table}

Our primary interest is the more fundamental generation bias. Each LLM generates text with its own model-specific patterns---characteristic n-gram distributions and stylistic tendencies. These patterns advantage the same model in two ways: (1)~the model translates text matching its own patterns more accurately, and (2)~since LLMs can recognize their own outputs~\citep{llm-favors-its-own-outputs}, they score them more favorably as evaluators. In high-resource languages, all models produce diverse text and these model-specific patterns are minimal. In low-resource languages, each model falls back on a narrow set of patterns, making its outputs distinctive and amplifying self-bias. To isolate generation bias from the translatability confound, we study source-only generation in subsequent sections.

\section{How does source text generation impact self-bias?}
\label{sec:ref_free_self_bias}

This section traces self-bias to the properties of generated source texts. We show that src-only generation still exhibits self-bias (\Cref{sec:self_bias_src_only}), that each model's implicit stylistic tendencies produce homogeneous outputs despite explicit diversity controls (\Cref{sec:pitall_src_only_gen}), that this lack of diversity drives self-bias (\Cref{sec:lack_diversity_quality}), and that improving diversity partially mitigates it (\Cref{sec:mitigate_self_bias}). We focus on Aymara$\rightarrow$English and Bemba$\rightarrow$English.

\subsection{Source text generation leads to self-bias}
\label{sec:self_bias_src_only}
As \Cref{tab:self_across_en_xx} shows, a measurable self-bias persists even when the data generation pipeline is constrained to produce only source texts without co-generated references. The negative bias scores appear exclusively on the diagonal, confirming that self-bias is inherent to each model's generated texts. Source-only generation thus does not eliminate self-bias; the model's inherent stylistic tendencies in monolingual generation are sufficient to create evaluation advantages for its own translation system.

\begin{table}[t]
\centering
\small
\begin{tabular}{l@{\hspace{1mm}}lm{11mm}m{11mm}m{11mm}}
& Bemba${\rightarrow}$En
& \multicolumn{3}{c}{LLM-as-a-benchmark} \\
&& \textbf{Gemini} & \textbf{GPT} & \textbf{Claude} \\
\cmidrule{2-5}
\parbox[t]{2mm}{\multirow{3}{*}{\rotatebox[origin=r]{90}{Translator\hspace{-5mm}}}}
& \bf Gemini & \cellcolor{Firebrick3!58}{-}0.515$^*$ & \cellcolor{SeaGreen3!13}\phantom{-}0.055 & \cellcolor{SeaGreen3!34}\phantom{-}0.460 \\
& \bf GPT & \cellcolor{SeaGreen3!27}\phantom{-}0.235 & \cellcolor{Firebrick3!26}{-}0.253$^*$ & \cellcolor{SeaGreen3!27}\phantom{-}0.018 \\
& \bf Claude & \cellcolor{SeaGreen3!8}\phantom{-}0.220 & \cellcolor{SeaGreen3!40}\phantom{-}0.543 & \cellcolor{Firebrick3!60}{-}0.763$^*$ \\
\bottomrule
\end{tabular}
\quad
\begin{tabular}{l@{\hspace{1mm}}lm{11mm}m{11mm}m{11mm}}
% \toprule
& Aymara${\rightarrow}$En
& \multicolumn{3}{c}{LLM-as-a-benchmark} \\
&& \textbf{Gemini} & \textbf{GPT} & \textbf{Claude} \\
\cmidrule{2-5}
\parbox[t]{2mm}{\multirow{3}{*}{\rotatebox[origin=r]{90}{Translator\hspace{-5mm}}}}
& \bf Gemini & \cellcolor{Firebrick3!23}{-}0.293$^*$ & \cellcolor{SeaGreen3!14}\phantom{-}0.285 & \cellcolor{SeaGreen3!7}\phantom{-}0.008 \\
& \bf GPT & \cellcolor{SeaGreen3!18}\phantom{-}0.308 & \cellcolor{Firebrick3!36}{-}0.330$^*$ & \cellcolor{SeaGreen3!7}\phantom{-}0.022 \\
& \bf Claude & \cellcolor{SeaGreen3!20}\phantom{-}0.015 & \cellcolor{SeaGreen3!29}\phantom{-}0.473 & \cellcolor{Firebrick3!60}{-}0.488$^*$ \\
\bottomrule
\end{tabular}
\caption{\llmbench self-bias with src-only generation for Bemba$\rightarrow$En and Aymara$\rightarrow$En. Diagonal entries (self-bias) are significantly more negative than peer ranking bias ($^*p<0.05$) for all models, confirming that self-bias persists even without co-generated references.}
\label{tab:self_across_en_xx}
\end{table}

\subsection{What makes source text generation exhibit bias?}
\label{sec:pitall_src_only_gen}

We investigate how each LLM's source text characteristics contribute to self-bias. Each model's source texts are more similar to its own other generated texts---even those for entirely different topics---than to texts from other models. To quantify this within-model versus cross-model similarity, we define a chrF@K metric.

Let $M = \{M_1, M_2, M_3\}$ be the set of LLMs. Each model $M_k \in M$ generates $N$ source texts $S_k = \{s_{k,1}, \ldots, s_{k,N}\}$, where the index $j$ in $s_{k,j}$ denotes the topic ID. For a source text $s_{A,i} \in S_A$, we compute pairwise chrF scores against all $s_{B,j} \in S_B$ (excluding self-matches when $M_A{=}M_B$), select the $K{=}5$ highest scores, and average them:
\begin{equation}
\text{chrF@K}(s_{A,i}, M_B) = \frac{1}{K} \sum_{k=1}^{K} \text{chrF}_k
\end{equation}
Averaging over all $s_{A,i} \in S_A$ yields the model-level similarity:
\begin{equation}
\text{chrF@K}(M_A, M_B) = \frac{1}{N} \sum_{i=1}^{N} \text{chrF@K}(s_{A,i}, M_B)
\label{eqn:similarity}
\end{equation}
This process yields three scores for each model $M_A$: one for its average within-model similarity (diagonal, $M_A{=}M_B$), and two for its cross-model similarities to the other LLMs (off-diagonal). \Cref{tab:biased_chrf_table} presents the results. The diagonal entries---representing within-model similarity---are consistently higher than off-diagonal entries across all models and language pairs. This demonstrates that each model exhibits significantly higher textual similarity to its own generated content, even across different topics, compared to texts generated by other models. Rather than producing truly diverse source texts following the prompt instructions, the models tend to reproduce content and style patterns from their limited knowledge, especially for low-resource languages like Bemba and Aymara.

\begin{table}[t]
\centering
\small
\begin{tabular}{l@{\hspace{1mm}}lp{1cm}p{1cm}p{1cm}}
% \toprule
& Bemba${\rightarrow}$En
& \multicolumn{3}{c}{chrF across prompts + models} \\
&& \textbf{Gemini} & \textbf{GPT} & \textbf{Claude} \\
\cmidrule{2-5}
\parbox[t]{2mm}{\multirow{3}{*}{\rotatebox[origin=r]{90}{Data from\hspace{-5mm}}}}
& \bf Gemini & \cellcolor{Firebrick3!60}35.76* & \cellcolor{SeaGreen3!30}32.70 & \cellcolor{SeaGreen3!25}32.03 \\
& \bf GPT & \cellcolor{SeaGreen3!27}32.28 & \cellcolor{Firebrick3!60}37.78* & \cellcolor{SeaGreen3!20}31.63 \\
& \bf Claude & \cellcolor{SeaGreen3!35}36.42 & \cellcolor{SeaGreen3!34}35.99 & \cellcolor{Firebrick3!60}38.64* \\
\bottomrule
\end{tabular}
\quad
\begin{tabular}{l@{\hspace{1mm}}lp{1cm}p{1cm}p{1cm}}
& Aymara${\rightarrow}$En
& \multicolumn{3}{c}{chrF across prompts + models} \\
&& \textbf{Gemini} & \textbf{GPT} & \textbf{Claude} \\
\cmidrule{2-5}
\parbox[t]{2mm}{\multirow{3}{*}{\rotatebox[origin=r]{90}{Data from\hspace{-5mm}}}}
& \bf Gemini & \cellcolor{Firebrick3!60}37.78* & \cellcolor{SeaGreen3!22}32.19 & \cellcolor{SeaGreen3!25}32.58 \\
& \bf GPT & \cellcolor{SeaGreen3!33}33.52 & \cellcolor{Firebrick3!60}40.79* & \cellcolor{SeaGreen3!24}32.44 \\
& \bf Claude & \cellcolor{SeaGreen3!38}39.56 & \cellcolor{SeaGreen3!37}37.04 & \cellcolor{Firebrick3!60}42.28* \\
\bottomrule
\end{tabular}
\caption{Average chrF@K similarity scores. Diagonal (within-model) entries are statistically higher than off-diagonal (cross-model) entries ($^*p<0.05$), showing each model's outputs are more similar to itself than to other models.}
\label{tab:biased_chrf_table}
\end{table}

\subsection{Limited Diversity in Low-Resource Source Texts}
\label{sec:lack_diversity_quality}
We hypothesize that the elevated within-model similarity stems from limited lexical diversity in low-resource language generation. We measure two complementary metrics: the Type-Token Ratio (TTR), which captures the proportion of unique words in a text and serves as a standard measure of lexical diversity, and the degeneration ratio, which quantifies the frequency of pathological repetition in generated outputs.

\begin{figure}[t]
    \centering
    \includegraphics[width=0.95\linewidth]{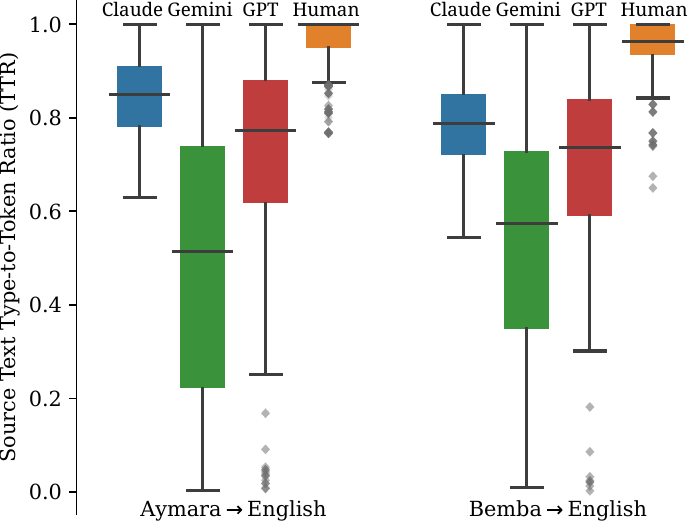}
    \vspace{1mm}
    {\small
    \begin{tabular}{lr}
    \toprule
    \bf Model & \bf Degeneration~(\%) \\
    \midrule
    Gemini & 22.3 \\
    GPT4.1 & 5.8  \\
    Claude & 0    \\
    \bottomrule
    \end{tabular}}
    \caption{TTR of LLM-generated vs.~human (FLORES) source texts. All LLMs are less lexically diverse than humans, and each model exhibits a distinct TTR profile. Degeneration (\% texts with $\geq$10 repeating 4-grams) is worst for Gemini (22.3\%).}
    \label{fig:ttr_and_stats_xx_en}
\end{figure}

\Cref{fig:ttr_and_stats_xx_en} illustrates the TTR distributions and degeneration ratios across the three models. Each model exhibits a visually distinct TTR profile, confirming that their lexical diversity patterns diverge substantially. Notably, all three LLMs---including Claude, which shows the highest diversity---are less lexically diverse than human-written benchmarks\footnote{We use $200$ source texts from FLORES-200 Aymara and Bemba as the human baseline.}, supporting prior work~\citep{yu2023largelanguagemodelattributed} on data diversity's role in mitigating systematic bias.

Further inspection reveals that the notably low diversity in Gemini-2.5-pro and GPT-4.1 outputs often stems from text degeneration. \textbf{We quantify degeneration by counting repeating n-grams, marking texts with $\geq$10 repeating 4-grams as degenerated.} As shown in \Cref{fig:ttr_and_stats_xx_en}, both models exhibit substantial degeneration levels (22.3\% for Gemini, 5.8\% for GPT-4.1), while Claude shows none. Crucially, degeneration can be a direct driver of self-bias: when a model translates its own degenerated source texts, translation quality improves compared to when other models translate the same inputs (see Appendix \Cref{sec:self_repair_degenerated}). This asymmetry---where the generating model handles its own degeneration artifacts better than its competitors---inflates its translation scores on its own test set, which contributes to the self-bias we observe.

\subsection{Improving diversity reduces self-bias}
\label{sec:mitigate_self_bias}

We hypothesize that the lack of diversity in generated source texts is a key contributor to self-bias, and test this by conducting an ablation study using our within-model similarity metric. As defined in \Cref{eqn:similarity} (with $M_A{=}M_B$), a high within-model similarity score indicates a model's tendency to repeat its own content patterns. For each of the three LLMs, we selected three subsets of $50$ source texts from the total $200$: those with the highest within-model chrF@K similarity (representing the least lexically diverse texts), those with the lowest similarity (representing the most diverse texts), and a randomly selected subset as a control. In preliminary experiments, we found that $K$ is not a sensitive parameter (tested $K{=}3$ to $7$), and that generating $4{\times}$ the target data provides sufficient room for diversity filtering.

To isolate the effect of source text diversity from evaluator bias, we evaluate these subsets under \llmtest only, where translation quality is measured by MetricX-QE. As shown in \Cref{tab:diversity_self_bias_testset}, Min chrF achieves significantly lower self-bias than Max chrF for all three models ($p<0.05$), confirming that \textbf{more diverse source texts reduce testset-generation bias}. The effectiveness varies across models: the reduction is strongest for Gemini and GPT, and more modest for Claude. Full \llmbench results (where evaluator bias also contributes) appear in Appendix~\Cref{sec:diversity_benchmark_appendix}.

\begin{table}[t]
\centering
\small
    \begin{tabular}{lm{10mm}m{10mm}m{10mm}}
    \toprule
    \llmtest & \bf Gemini & \bf GPT & \bf Claude \\
    \midrule
    \bf Max chrF & {-}0.240$^*$ & {-}0.296$^*$ & {-}0.205$^*$ \\
    \bf Random  & {-}0.207$^*$ & {-}0.240$^*$ & {-}0.053 \\
    \bf Min chrF  & {-}0.170 & {-}0.195 & \phantom{-}0.045 \\
    \bottomrule
    \end{tabular}
\caption{Self-bias of \llmtest for subsets of source texts selected by within-model chrF@K similarity. Min chrF (most diverse) achieves significantly lower self-bias than Max chrF (least diverse) for all models ($^*p<0.05$). Self-bias is averaged for Aymara$\rightarrow$En and Bemba$\rightarrow$En.}
\label{tab:diversity_self_bias_testset}
\end{table}

\section{Self-Bias is Associated with Language Proficiency}
\label{sec:language_proficiency}

Self-bias stems from limited diversity in model-generated texts---but \textit{when does this diversity collapse, and what governs its severity?} We show that language proficiency is a key contributing factor. When models generate text in languages where they are highly proficient (e.g., English), their outputs are lexically varied and stylistically diverse across models, substantially reducing model-specific signatures and, consequently, self-bias. Conversely, in low-proficiency languages, models default to a narrow set of patterns, amplifying self-bias. We demonstrate this through translation direction asymmetry (\Cref{sec:translation_direction}) and cross-task validation on open-ended generation via Chatbot Arena (\Cref{sec:non_translation}).

\subsection{Translation Direction Asymmetry}
\label{sec:translation_direction}

\Cref{tab:ablation_into_out_of_english} shows a striking asymmetry: En${\rightarrow}$XX directions exhibit substantially lower self-bias than XX${\rightarrow}$En. The magnitude of self-bias for English${\rightarrow}$Aymara and English${\rightarrow}$Bemba is consistently less than $0.15$ in absolute value, whereas the XX${\rightarrow}$En directions show much larger effects. \Cref{tab:ablation_into_out_of_english_testset_evaluator} decomposes \llmbench into its \llmtest and \llmeval components by direction. Both components consistently exhibit more pronounced self-bias in the XX${\rightarrow}$En direction than in En${\rightarrow}$XX. This suggests that when generating source texts in low-resource languages, the \llmtest produces texts with stronger model-specific linguistic features that advantage its own translation system. Simultaneously, the \llmeval appears more attuned to these self-generated patterns when judging XX${\rightarrow}$En outputs, leading to systematically inflated scores.

\begin{table}[t]
    \centering
    \small
    \setlength{\tabcolsep}{3pt}
    \begin{tabular}{m{1.2cm}m{1.8cm}m{11mm}m{11mm}m{11mm}}
    \toprule
    \bf Lang & \bf Direction & \bf Gemini & \bf GPT & \bf Claude \\
    \midrule
    \multirow{2}{*}{Bemba}
    & En${\rightarrow}$Bemba & \cellcolor{Firebrick3!19}{-}0.15 & \cellcolor{Firebrick3!0}\phantom{-}0.10 & \cellcolor{Firebrick3!0}\phantom{-}0.14 \\
    & Bemba${\rightarrow}$En & \cellcolor{Firebrick3!77}{-}0.59$^*$ & \cellcolor{Firebrick3!26}{-}0.20$^*$ & \cellcolor{Firebrick3!80}{-}0.62$^*$ \\
    \midrule
    \multirow{2}{*}{Aymara}
    & En${\rightarrow}$Aymara & \cellcolor{Firebrick3!0}\phantom{-}0.02 & \cellcolor{Firebrick3!12}{-}0.09 & \cellcolor{Firebrick3!0}\phantom{-}0.07 \\
    & Aymara${\rightarrow}$En & \cellcolor{Firebrick3!20}{-}0.16$^*$ & \cellcolor{Firebrick3!41}{-}0.32$^*$ & \cellcolor{Firebrick3!80}{-}0.62$^*$ \\
    \bottomrule
    \end{tabular}
    \caption{\llmbench self-bias by translation direction. XX$\rightarrow$En directions show significantly more self-bias than En$\rightarrow$XX for all models ($^*p<0.05$, permutation test).}
    \label{tab:ablation_into_out_of_english}
\end{table}

\begin{table}[t]
    \centering
    \small
    \setlength{\tabcolsep}{3pt}
    \begin{tabular}{m{1.8cm}m{1.2cm}m{11mm}m{11mm}m{11mm}}
    \toprule
    \bf Setting & \bf Direction & \bf Gemini & \bf GPT & \bf Claude \\
    \midrule
    \multirow{2}{*}{As-a-testset}
    & En${\rightarrow}$XX & \cellcolor{SeaGreen3!0}\phantom{-}0.02 & \cellcolor{SeaGreen3!0}\phantom{-}0.05 & \cellcolor{Firebrick3!14}{-}0.08 \\
    & XX${\rightarrow}$En & \cellcolor{Firebrick3!32}{-}0.17$^*$ & \cellcolor{Firebrick3!43}{-}0.24$^*$ & \cellcolor{Firebrick3!17}{-}0.09 \\
    \midrule
    \multirow{2}{*}{As-an-evaluator}
    & En${\rightarrow}$XX & \cellcolor{Firebrick3!20}{-}0.11 & \cellcolor{SeaGreen3!0}\phantom{-}0.05 & \cellcolor{SeaGreen3!0}\phantom{-}0.10 \\
    & XX${\rightarrow}$En & \cellcolor{Firebrick3!54}{-}0.30$^*$ & \cellcolor{Firebrick3!80}{-}0.44$^*$ & \cellcolor{Firebrick3!55}{-}0.30$^*$ \\
    \bottomrule
    \end{tabular}
    \caption{Self-bias decomposed into \llmtest and \llmeval by direction. XX$\rightarrow$En shows significantly stronger bias than En$\rightarrow$XX ($^*p<0.05$), except Claude under \llmtest.}
    \label{tab:ablation_into_out_of_english_testset_evaluator}
\end{table}

\paragraph{Why does translation asymmetry exist for self-bias?}
\Cref{tab:tab_ttr_en_xx_xx_en_comparisons} reports pairwise Cohen's D~\citep{cohen1988statistical} between models' TTR distributions, measuring how much each pair of models diverges in their lexical diversity profiles ($D{>}0.5$ indicates substantial divergence; see Appendix~\Cref{sec:cohen_d}).

\begin{table}[t]
\centering
\small
\setlength{\tabcolsep}{3pt}
\begin{tabular}{lrrr}
\toprule
 & \textbf{Gem.${\times}$Claude} & \textbf{Gem.${\times}$GPT} & \textbf{Claude${\times}$GPT} \\
\midrule
English${\rightarrow}$XX\hspace{-2mm} & 0.076 & 0.190 & 0.111 \\
XX${\rightarrow}$English\hspace{-2mm} & \textbf{1.487} & \textbf{0.777} & \textbf{0.663} \\
\bottomrule
\end{tabular}

% \begin{tabular}{lrrrrrr}
% \toprule
%  & \multicolumn{6}{c}{MetricX distribution differences on translations of human and ZSB sources at src paired with ref} \\
%  & \multicolumn{3}{c}{Bemba${\rightarrow}$En} & \multicolumn{3}{c}{Aymara${\rightarrow}$En} \\
% \cmidrule(r){2-4} \cmidrule(l){5-7}
% Lang dir & \textbf{Gem${\times}$Cla} & \textbf{Gem${\times}$GPT} & \textbf{Cla${\times}$GPT} & \textbf{Gem${\times}$Cla} & \textbf{Gem${\times}$GPT} & \textbf{Cla${\times}$GPT} \\
% \midrule
% ZSB & 0.036 & 0.078 & 0.117 & 0.097 & 0.027 & 0.130\\ 
% Human & \textbf{0.185} & \textbf{0.081} & \textbf{0.268} & \textbf{0.390} & \textbf{0.284} & \textbf{0.152}\\
% \bottomrule
% \end{tabular}

\caption{Lexical diversity divergence across models (pairwise TTR Cohen's D). English source texts show near-zero divergence, while low-resource source texts diverge substantially ($D{>}0.5$).}
\label{tab:tab_ttr_en_xx_xx_en_comparisons}
\end{table}

\textbf{En$\rightarrow$XX (low self-bias):} When generating English source texts, all three models are highly proficient and follow the diversity prompt effectively. As a result, all three produce lexically diverse texts---their TTR distributions are independently high and look similar across models, yielding low pairwise Cohen's D ($D{<}0.2$ for all pairs). Importantly, this does \emph{not} mean the models produce the same texts; it means that none of the models leave distinctive low-diversity patterns. Without model-specific signatures, translators and evaluators are less likely to favor their own sources over competitors', which is consistent with the suppressed self-bias we observe.

\textbf{XX$\rightarrow$En (high self-bias):} When generating low-resource language texts, limited proficiency causes each model to fall back on its own narrow, repetitive patterns---model-specific artifacts. Each model's outputs become low in TTR, but the specific degeneration patterns differ across models (e.g., Gemini degenerates 22.3\% of texts, Claude 0\%). This creates \emph{divergent} low-diversity TTR distributions, reflected in high pairwise Cohen's D ($D{>}0.5$ for all pairs). These distinctive generation patterns plausibly give the generating model a systematic advantage: its translator handles its own artifacts better than other models' translators do, and its evaluator may recognize its own stylistic patterns more favorably. This two-sided advantage is consistent with the large self-bias we observe for XX$\rightarrow$En. Correspondingly, Appendix \cref{tab:biased_chrf_table_en_xx} shows that within-model chrF similarity is not elevated for English sources, whereas XX$\rightarrow$En diagonal entries are consistently higher, confirming the presence of model-specific generation patterns only in the low-resource direction.

\subsection{Cross-Task Validation: Chatbot Arena}
\label{sec:non_translation}
To confirm generality, we apply \llmbench to the Chatbot Arena task~\citep{chiang2024chatbotarenaopenplatform} using prompts from \citet{pombal2025zeroshot} (details in Appendix~\Cref{sec:prompt_setup}), comparing high-resource (English, Korean) versus low-resource (Bemba, Aymara) settings.

As shown in \cref{tab:llmarena_results}, results closely mirror our translation findings. Self-bias is near-zero in high-resource settings but increases sharply for low-resource prompts. \textbf{This confirms that self-bias associated with low LLM proficiency generalizes beyond translation.}

\begin{table}[t]
\centering
\small
\begin{tabular}{l@{\hspace{1mm}}lp{1cm}p{1cm}p{1cm}}
% \toprule
& English
& \multicolumn{3}{c}{LLM-as-a-benchmark} \\
&& \textbf{Gemini} & \textbf{GPT} & \textbf{Claude} \\
\cmidrule{2-5}
\parbox[t]{2mm}{\multirow{3}{*}{\rotatebox[origin=r]{90}{Generator\hspace{-5mm}}}}
& \bf Gemini & \cellcolor{Red3!0}-0.018 & \cellcolor{Red3!15}-0.122 & \cellcolor{SeaGreen3!40}0.140 \\
& \bf GPT & \cellcolor{SeaGreen3!15}0.035 & \cellcolor{Red3!40}-0.205 & \cellcolor{SeaGreen3!50}0.170 \\
& \bf Claude & \cellcolor{SeaGreen3!30}0.103 & \cellcolor{Red3!15}-0.107 & \cellcolor{SeaGreen3!5}0.005 \\
\bottomrule
\end{tabular}
\quad
\begin{tabular}{l@{\hspace{1mm}}lp{1cm}p{1cm}p{1cm}}
& Korean
& \multicolumn{3}{c}{LLM-as-a-benchmark} \\
&& \textbf{Gemini} & \textbf{GPT} & \textbf{Claude} \\
\cmidrule{2-5}
\parbox[t]{2mm}{\multirow{3}{*}{\rotatebox[origin=r]{90}{Generator\hspace{-5mm}}}}
& \bf Gemini & \cellcolor{SeaGreen3!5}0.068 & \cellcolor{Red3!5}-0.060 & \cellcolor{Red3!5}-0.007 \\
& \bf GPT & \cellcolor{SeaGreen3!35}0.107 & \cellcolor{Red3!40}-0.208 & \cellcolor{SeaGreen3!30}0.100 \\
& \bf Claude & \cellcolor{SeaGreen3!35}0.105 & \cellcolor{Red3!5}-0.068 & \cellcolor{Red3!5}-0.038 \\
\bottomrule
\end{tabular}

\vspace{2mm}

\begin{tabular}{l@{\hspace{1mm}}lp{1cm}p{1cm}p{1cm}}
% \toprule
& Bemba
& \multicolumn{3}{c}{LLM-as-a-benchmark} \\
&& \textbf{Gemini} & \textbf{GPT} & \textbf{Claude} \\
\cmidrule{2-5}
\parbox[t]{2mm}{\multirow{3}{*}{\rotatebox[origin=r]{90}{Generator\hspace{-5mm}}}}
& \bf Gemini & \cellcolor{Red3!50}-0.385 & \cellcolor{Red3!5}-0.010 & \cellcolor{SeaGreen3!65}0.396 \\
& \bf GPT & \cellcolor{SeaGreen3!15}0.052 & \cellcolor{Red3!45}-0.218 & \cellcolor{SeaGreen3!35}0.167 \\
& \bf Claude & \cellcolor{SeaGreen3!60}0.334 & \cellcolor{Red3!18}-0.154 & \cellcolor{Red3!40}-0.180 \\
\bottomrule
\end{tabular}
\quad
\begin{tabular}{l@{\hspace{1mm}}lp{1cm}p{1cm}p{1cm}}
& Aymara
& \multicolumn{3}{c}{LLM-as-a-benchmark} \\
&& \textbf{Gemini} & \textbf{GPT} & \textbf{Claude} \\
\cmidrule{2-5}
\parbox[t]{2mm}{\multirow{3}{*}{\rotatebox[origin=r]{90}{Generator\hspace{-5mm}}}}
& \bf Gemini & \cellcolor{Red3!45}-0.216 & \cellcolor{SeaGreen3!60}0.346 & \cellcolor{Red3!15}-0.130 \\
& \bf GPT & \cellcolor{SeaGreen3!45}0.247 & \cellcolor{Red3!50}-0.413 & \cellcolor{SeaGreen3!35}0.165 \\
& \bf Claude & \cellcolor{SeaGreen3!40}0.177 & \cellcolor{Red3!25}-0.108 & \cellcolor{Red3!15}-0.069 \\
\bottomrule
\end{tabular}
\caption{\llmbench self-bias on Chatbot Arena. Low-resource prompts (Bemba, Aymara) elicit much stronger self-bias than high-resource prompts (English, Korean).}
\label{tab:llmarena_results}
\end{table}

\section{Merits of \llmbench}

Despite self-bias concerns, \llmbench retains practical value in two settings.

\paragraph{\llmbench benefits open-source model evaluation.} \Cref{tab:bias_open_src_models} shows that frontier LLMs consistently rank open-source models (Gemma3-27B, Mistral-large-2411, Qwen3-32B) with low bias---except GPT-4.1's specific bias toward Qwen3-32B. When the quality gap between generator and evaluated models is large, self-bias does not distort rankings. However, bias remains substantial when frontier models rank each other.

\paragraph{High-resource languages reduce self-bias.} Self-bias is substantially reduced for high-resource language pairs (\Cref{sec:language_proficiency}), and applying diversity filtering to source texts mitigates it further (\Cref{sec:mitigate_self_bias}). Practitioners should exercise particular caution when using \llmbench for low-resource languages or when comparing models of similar capability.

\begin{table}[t]
\centering
\small
\begin{tabular}{l@{\hspace{1mm}}lp{1cm}p{1cm}p{1cm}}
\toprule
& Bemba
& \multicolumn{3}{c}{LLM-as-a-benchmark (Rank)} \\
&& \textbf{Gemini} & \textbf{GPT} & \textbf{Claude} \\
\cmidrule{2-5}
\parbox[t]{3mm}{\multirow{3}{*}{\rotatebox[origin=r]{90}{Translator\hspace{-5mm}}}}
& \bf Gemma3 & \cellcolor{Firebrick3!58}1.125 & \cellcolor{Firebrick3!58}1.110 & \cellcolor{Firebrick3!58}1.160 \\
& \bf Mistral & \cellcolor{SeaGreen3!27}1.675 & \cellcolor{SeaGreen3!27}1.623 & \cellcolor{SeaGreen3!27}1.630 \\
& \bf Qwen3 & \cellcolor{SeaGreen3!50}2.335 & \cellcolor{SeaGreen3!58}2.558 & \cellcolor{SeaGreen3!48}2.285 \\
\bottomrule % Bottom rule for a cleaner look, requires booktabs
\end{tabular}
\quad
\begin{tabular}{l@{\hspace{1mm}}lp{1cm}p{1cm}p{1cm}}
\toprule
& Bemba
& \multicolumn{3}{c}{LLM-as-a-benchmark (Bias)} \\
&& \textbf{Gemini} & \textbf{GPT} & \textbf{Claude} \\
\cmidrule{2-5}
\parbox[t]{3mm}{\multirow{3}{*}{\rotatebox[origin=r]{90}{Translator\hspace{-5mm}}}}
& \bf Gemma3 & \cellcolor{SeaGreen3!8}-0.01 & \cellcolor{SeaGreen3!8}\phantom{-}0.04 & \cellcolor{SeaGreen3!8}-0.03 \\
& \bf Mistral & \cellcolor{SeaGreen3!8}\phantom{-}0.05 & \cellcolor{SeaGreen3!8}-0.02 & \cellcolor{SeaGreen3!8}-0.03 \\
& \bf Qwen3 & \cellcolor{SeaGreen3!8}-0.09 & \cellcolor{Firebrick3!18}-0.16 & \cellcolor{SeaGreen3!48}\phantom{-}0.25 \\
\bottomrule
\end{tabular}
\caption{Frontier models rank open-source models (Gemma3-27B, Mistral-large-2411, Qwen3-32B) consistently and with low bias~\citep{pombal2025zeroshot}, except GPT-4.1's bias toward Qwen3-32B. Self-bias remains significant when frontier models rank each other.}
\label{tab:bias_open_src_models}
\end{table}

\section{Related Work}

\paragraph{Automatic Benchmark Creation.} As existing benchmarks become increasingly saturated by rapid LLM advancements~\cite{glazer2025frontiermathbenchmarkevaluatingadvanced}, automatic benchmark construction using LLMs has gained significant traction~\cite{pombal2025zeroshotbenchmarkingframeworkflexible}. Methods range from synthesizing test sets via prompt workflows~\citep{sprague2024musrtestinglimitschainofthought,zouhar-etal-2025-generating} to constructing prompts from existing data~\citep{li2024crowdsourceddatahighqualitybenchmarks,vergopoulos2025automated}. Efficacy is assessed via benchmark agreement~\citep{perlitz2024llmbenchmarksagreefixing} or ranking correlation with human benchmarks~\citep{pombal2025zeroshotbenchmarkingframeworkflexible}, using accuracy for verifiable tasks or LLM judges for open-ended outputs~\citep{xu-etal-2023-instructscore, pombal2025zeroshotbenchmarkingframeworkflexible}. A key advantage is the ability to generate evaluation data for arbitrary domains and language pairs on demand, which is particularly valuable for settings where human data curation is expensive or difficult.

\paragraph{Self-bias in LLM Evaluation.} The \llmbench paradigm is susceptible to self-bias from two sources: the LLM acting as an evaluator and the LLM generating the test set. A well-documented issue is the tendency of LLM judges to systematically favor their own outputs~\citep{xu-etal-2024-pride, llm-favors-its-own-outputs}, attributed to familiarity with their own stylistic patterns or a preference for low-perplexity text~\citep{wataoka2025selfpreferencebiasllmasajudge}. While this preference may sometimes reflect genuine quality~\citep{chen2025llmevaluatorspreferreason}, judge reliability is questionable for problems beyond the model's own capability~\citep{krumdick2025freelabelslimitationsllmasajudge}. Our work extends this line of research by investigating the overlooked self-bias introduced by testset generation---an independent source of bias absent from prior setups that hold test inputs constant---and showing that the two sources of bias compound when combined. Unlike prior work that studies evaluator self-preference on human-authored test data, we demonstrate that testset generation alone can flip model rankings. Unlike \citet{yuan2025silencerdiscoverymitigationselfbias}, whose mitigation strategies target verifiable tasks like math reasoning, we address generative settings where evaluation itself requires subjective judgment.

\section{Conclusion}

We formally define and quantify self-bias in \llmbench, decomposing it into compounding \llmtest and \llmeval components. Using machine translation as our primary testbed, we demonstrate that self-bias causes each model to rank itself first relative to peer consensus, and validate this on the Chatbot Arena task. We identify limited model proficiency as a plausible driver: in low-resource settings, each model's implicit tendencies produce homogeneous, model-specific outputs whose degeneration patterns the generating model handles better than its competitors.

Based on these findings, we offer actionable guidance: increasing source text diversity using our proposed diversity metric reduces self-bias, though the effectiveness varies across models; generating benchmark data in high-resource languages where models exhibit similarly high diversity minimizes model-specific patterns; and using frontier models to benchmark open-source models remains reliable when the quality gap is large.

A promising future direction is applying model obfuscation techniques---such as paraphrasing generated texts with a different model---to remove model-specific patterns, though this risks introducing the rewriter's own biases. We hope this work alerts practitioners to self-bias as a validity threat in automated benchmarking and motivates the development of frameworks robust to models' generative idiosyncrasies.

{
\color{blue}
% \begin{itemize}
%                           \item Wenda: add things
% \end{itemize}
}

\section*{Limitations}

Our conclusions are based on three frontier closed-source LLMs and two tasks (machine translation and Chatbot Arena). While broader coverage of models, tasks, and languages would further strengthen generalizability, the consistent diagonal pattern across all tested configurations---every model ranks itself first---suggests the phenomenon is systematic rather than model-specific.

\paragraph{Peer consensus as proxy.} Our self-bias framework uses peer-consensus ranking as a proxy for true quality, derived from two peer models per judgment. While human rankings would provide stronger grounding, peer consensus is a good proxy in settings where human evaluation is infeasible---precisely the low-resource scenarios where \llmbench is most needed. Moreover, for \llmtest, we validate with MetricX-QE, an independent neural metric entirely outside the evaluated model families, and self-bias persists, confirming that our findings are not artifacts of the peer-consensus framework. As future work, large-scale human evaluations covering these models and language pairs would provide additional validation of our results and help disentangle self-preference from genuine quality differences.

\paragraph{Correlational vs.\ causal evidence.} Our diversity-filtering ablation demonstrates that reducing within-model similarity consistently reduces self-bias, providing interventional (not merely observational) evidence for the role of text diversity. A fully controlled experiment---e.g., paraphrasing all source texts to equalize diversity---would strengthen causal claims but risks introducing confounds from the paraphrasing model itself. We leave such controlled designs to future work.

\paragraph{Model-dependent mitigation.} Diversity filtering significantly reduces self-bias under \llmtest for all three models. Under full \llmbench, the effect is significant for Gemini and Claude but not for GPT, suggesting that evaluator-side bias may dominate for some models. This variation motivates future work on model-specific or combined mitigation strategies.
\bibliography{main,anthology.min}

%%%%%%%%%%%%%%%%%%%%%%%%%%%%%%%%%%%%%%%%%%%%%%%%%%%%%%%%%%%%%%%%%%%%%%%%%%%%%%%
%%%%%%%%%%%%%%%%%%%%%%%%%%%%%%%%%%%%%%%%%%%%%%%%%%%%%%%%%%%%%%%%%%%%%%%%%%%%%%%
% APPENDIX
%%%%%%%%%%%%%%%%%%%%%%%%%%%%%%%%%%%%%%%%%%%%%%%%%%%%%%%%%%%%%%%%%%%%%%%%%%%%%%%
%%%%%%%%%%%%%%%%%%%%%%%%%%%%%%%%%%%%%%%%%%%%%%%%%%%%%%%%%%%%%%%%%%%%%%%%%%%%%%%
\newpage
\appendix

\section{Statistical Significance and Sample Size Justification}
\label{sec:bootstrap}

To justify our sample size of $N{=}200$, we conduct a bootstrap analysis on the Bemba${\rightarrow}$English translation task under the \llmtest and \llmeval setup. For each of the three models (Gemini 2.5 Pro, GPT-4.1, Claude Opus 4), we construct the $3{\times}3$ ranking table and perform bootstrap resampling ($B{=}10{,}000$) to compute 95\% confidence intervals.

\Cref{tab:bootstrap_mean} reports the mean rankings with standard deviations, and \Cref{tab:bootstrap_ci} reports the corresponding 95\% confidence intervals. The diagonal entries---representing each model's ranking of itself---show no overlap with the off-diagonal entries in the same column, confirming that each model ranks itself significantly higher than the other two. Based on this analysis and budget constraints, we use $N{=}200$ for all experiments.

\begin{table}[h]
\centering
\small
\caption{Mean rankings ($\pm$ std) from bootstrap resampling on Bemba${\rightarrow}$English. Diagonal entries (self-rankings) are consistently lower (better) than off-diagonal entries.}
\label{tab:bootstrap_mean}
\begin{tabular}{lccc}
\toprule
\textbf{Model} & \textbf{Gemini} & \textbf{GPT-4.1} & \textbf{Claude} \\
\midrule
Gemini & $1.00 \pm 0.03$ & $1.52 \pm 0.09$ & $1.70 \pm 0.10$ \\
GPT-4.1 & $1.21 \pm 0.07$ & $1.18 \pm 0.05$ & $1.54 \pm 0.09$ \\
Claude & $1.64 \pm 0.11$ & $1.93 \pm 0.11$ & $1.17 \pm 0.06$ \\
\bottomrule
\end{tabular}
\end{table}

\begin{table}[h]
\centering
\small
\caption{95\% bootstrap confidence intervals for rankings on Bemba${\rightarrow}$English. Diagonal intervals do not overlap with off-diagonal intervals in each column, confirming statistical significance of self-bias.}
\label{tab:bootstrap_ci}
\begin{tabular}{lccc}
\toprule
\textbf{Model} & \textbf{Gemini} & \textbf{GPT-4.1} & \textbf{Claude} \\
\midrule
Gemini & $[0.97, 1.02]$ & $[1.43, 1.61]$ & $[1.59, 1.80]$ \\
GPT-4.1 & $[1.15, 1.28]$ & $[1.12, 1.23]$ & $[1.45, 1.64]$ \\
Claude & $[1.52, 1.75]$ & $[1.82, 2.04]$ & $[1.11, 1.23]$ \\
\bottomrule
\end{tabular}
\end{table}

\section{Cohen's d: Effect Size for Mean Differences} 
\label{sec:cohen_d}

Cohen's d \citep{cohen1988statistical} is a widely used standardized effect size measure to quantify the difference between two means, expressed in standard deviation units. It is particularly useful when comparing the central tendency of two groups (e.g., two sets of scores, two distributions) and interpreting the practical significance of their difference, independent of sample size.

Given two groups, Group 1 and Group 2, with means $\bar{x}_1$ and $\bar{x}_2$ and standard deviations $s_1$ and $s_2$ respectively, Cohen's d is typically calculated as:

$$ d = \frac{\bar{x}_1 - \bar{x}_2}{s_p} $$

where $s_p$ is the pooled standard deviation, calculated as:

$$ s_p = \sqrt{\frac{(n_1 - 1)s_1^2 + (n_2 - 1)s_2^2}{n_1 + n_2 - 2}} $$

$n_1$ and $n_2$ are the sample sizes for Group 1 and Group 2.

Interpretation Guidelines (Cohen's conventions):
\begin{itemize}
    \item $d = 0.2$: Small effect
    \item $d = 0.5$: Medium effect
    \item $d = 0.8$: Large effect
\end{itemize}

\newcommand{\promptexample}[2]{
\noindent \textbf{#1.}
{\fontsize{8.15pt}{8pt}\selectfont\tt #2 }
\medskip
}

\section{Prompts used for LLM-as-a-benchmark}
\label{sec:prompt_setup}
All prompts for both LLM-as-a-testset and LLM-as-an-evaluator are included. To ensure fair comparisons, these prompts are adapted from the zero-shot benchmark paper by \citep{pombal2025zeroshot}. The LLM-as-a-testset prompts incorporate a few randomly chosen seed variables—such as length, topic, subtopic, and style—to guarantee diverse generated source texts. A complete list of options for each seed variable is provided below.

\bigskip

\promptexample{Prompt for LLM-as-a-Testset (source + reference generation) for translation}{
You are a multilingual content creator and translation expert. Your task is to generate a comprehensive translation exercise based on the given attributes. Follow these instructions carefully:\\
1. Review the following input variables:\\
- Source language: {SOURCE LANGUAGE}\\
- Target language: {TARGET LANGUAGE}\\
- Topic: {topic}\\
- Subtopic: {subtopic}\\
- Source length: {length}\\
- Style: {style}\\
2. Generate a source text: Create an original text in the source language, adhering to the specified topic, subtopic, and length. The text should be coherent, informative, and suitable for translation.\\
3. Generate a reference translation: Produce a high-quality, fluent translation of the source text in the target language. \\
This translation should serve as a reference for evaluating other translations.
IT IS CRUCIAL THAT THE REFERENCE TRANSLATION SOUNDS NATURAL IN THE TARGET LANGUAGE.
Format your output as follows:\\
<START OF SOURCE>\\
INSERT THE SOURCE TEXT HERE\\
<END OF SOURCE>\\
<START OF REFERENCE TRANSLATION>\\
INSERT THE REFERENCE TRANSLATION HERE\\
<END OF REFERENCE TRANSLATION>\\
Ensure that your response is comprehensive, coherent, and follows all the instructions provided above. 
Abide strictly by the requested format and generated until the end of the requested output.
Only generate source and reference translation. Do not generate any other text such as reasoning or explanations.\\
<START OF SOURCE>
}

\bigskip

\promptexample{Prompt for LLM-as-a-Testset (source only) for translation}{
You are a multilingual content creator and translation expert. Your task is to generate a comprehensive translation exercise based on the given attributes. Follow these instructions carefully:\\
1. Review the following input variables:\\
- Source language: {SOURCE LANGUAGE}\\
- Topic: topic\\
- Subtopic: {subtopic}\\
- Source length: {length}\\
- Style: {style}\\
2. Generate a source text: Create an original text in the source language, adhering to the specified topic, subtopic, and length. The text should be coherent, informative, and suitable for translation.\\
Format your output as follows:\\
<START OF SOURCE>\\
INSERT THE SOURCE TEXT HERE\\
<END OF SOURCE>\\
Ensure that your response is comprehensive, coherent, and follows all the instructions provided above. \\
Abide strictly by the requested format and generated until the end of the requested output.
Only generate source text. Do not generate any other text such as reasoning or explanations.\\
<START OF SOURCE>
}

\promptexample{Topics}{
"Tech Innovation",
    "Global Markets",
    "Environmental Policy",
    "Public Health",
    "Urban Development",
    "International Relations",
    "Education Reform",
    "Cultural Trends",
    "Scientific Discoveries",
    "Economic Policy",
    "Sports Industry",
    "Media \& Entertainment",
    "Workplace Transformation",
    "Transportation \& Mobility",
    "Food \& Agriculture",
    "Medical \& Healthcare",
    "Legal \& Compliance",
    "E-commerce \& Retail",
    "Financial Services",
    "Gaming \& Software",
    "Marketing \& Advertising",
    "Government Documentation",
    "Academic Research",
    "Patents \& Intellectual Property",
    "Manufacturing \& Safety",
    "Tourism \& Hospitality",
    "Religious \& Cultural Studies",
    "Insurance \& Risk Management",
    "Consumer Electronics",
    "Pharmaceutical Industry",
    "Fashion \& Apparel",
    "Beauty \& Cosmetics",
    "Home \& Living",
    "Automotive Industry",
    "Social Media",
    "Dating \& Relationships",
    "Parenting \& Family",
    "Arts \& Culture",
    "Music Industry",
    "Film \& Cinema",
    "Books \& Literature",
    "Food \& Cuisine",
    "Sports \& Recreation",
    "Fitness \& Wellness",
    "Mental Health",
    "Architecture \& Design",
    "Real Estate",
    "Telecommunications",
    "Renewable Energy",
    "Space Exploration",
    "Wildlife \& Nature",
    "Weather \& Climate",
    "History \& Heritage",
    "Politics \& Governance",
    "NGOs \& Nonprofits",
    "New York City",
    "London",
    "Tokyo",
    "Paris",
    "Berlin",
    "Singapore",
    "Dubai",
    "São Paulo",
    "Sydney",
    "Mumbai",
    "Madrid",
    "Lisbon",
    "Stockholm",
    "Amsterdam",
    "Seoul",
    "Japan",
    "France",
    "Germany",
    "Brazil",
    "India",
    "Italy",
    "Spain",
    "China",
    "United Kingdom",
    "Portugal",
    "Poetry"
}

\promptexample{Subtopics}{
"Poetry": [
        "Modernism",
        "Contemporary",
        "Modernism",
        "Haiku",
        "European Poetry",
        "Asian Poetry",
        "Theme identification",
    ],
    "Tech Innovation": [
        "Artificial Intelligence",
        "Quantum Computing",
        "Robotics",
        "5G/6G Networks",
        "Biotechnology",
        "Green Tech",
        "Edge Computing",
        "Cybersecurity",
    ],
    "Global Markets": [
        "Stock Exchanges",
        "Cryptocurrency",
        "International Trade",
        "Foreign Investment",
        "Commodity Markets",
        "Emerging Markets",
        "Foreign Exchange",
        "Market Regulations",
    ],
    "Environmental Policy": [
        "Carbon Trading",
        "Renewable Energy Initiatives",
        "Wildlife Protection",
        "Urban Planning",
        "Waste Management",
        "Climate Agreements",
        "Marine Conservation",
    ],
    "Public Health": [
        "Disease Prevention",
        "Healthcare Systems",
        "Vaccination Programs",
        "Mental Health Services",
        "Maternal Health",
        "Epidemiology",
        "Health Technology",
    ],
    ......
}

\promptexample{Styles}{
"creative",
    "concise",
    "technical",
    "formal",
    "informal",
    "narrative",
    "persuasive",
    "descriptive",
    "analytical",
    "humorous",
    "poetic",
    "casual",
    "academic",
    "journalistic",
    "neutral",
    "elaborate",
    "minimalist",
    "rushed"
}

\promptexample{Lengths}{"short",
    "medium"}

\promptexample{Prompt for LLM-as-an-Evaluator  for translation}{You are an expert judge evaluating translation quality. You will be presented with:
- An original text
- A translation to evaluate

Rate the translation on a scale of 1-6 based on these key criteria:
- Accuracy and fidelity to source
- Grammar and language correctness
- Natural flow and readability
- Terminology consistency
- Completeness of translation
- Technical precision

Scoring Rubric:

6 - Outstanding
- Perfect accuracy with source meaning
- Flawless grammar and language use
- Reads completely naturally in target language
- Consistent and precise terminology
- Complete translation with no omissions
- Excellent technical accuracy

5 - Excellent
- Very accurate rendering of source
- Strong grammar with minimal issues
- Natural-sounding translation
- Good terminology consistency
- Nearly complete coverage
- Strong technical accuracy

4 - Good
- Generally accurate translation
- Mostly correct grammar
- Readable with some awkward passages
- Generally consistent terminology
- Minor omissions only
- Adequate technical accuracy

3 - Fair
- Some accuracy issues
- Notable grammar problems
- Often unnatural phrasing
- Inconsistent terminology
- Several omissions
- Technical inaccuracies present

2 - Poor
- Significant accuracy issues
- Frequent grammar errors
- Unnatural throughout
- Poor terminology consistency
- Major omissions
- Many technical errors

1 - Inadequate
- Fails to convey source meaning
- Severe grammar issues
- Incomprehensible in target language
- No terminology consistency
- Incomplete translation
- Technical meaning lost

Format your output as follows: 
Put detailed explanation between <START OF FEEDBACK> and </END OF FEEDBACK>
Put result between <START OF RESULT> and </END OF RESULT>
Don't provide any other text

<START OF FEEDBACK>
Put detailed explanation of the score based on the criteria here
</END OF FEEDBACK>\\\\
<START OF RESULT>
Put only a number from 1 to 6 here
</END OF RESULT>
<START OF SOURCE TEXT>
{prompt}
</END OF SOURCE TEXT>\\\\
<START OF TRANSLATION>
{answer}
</END OF TRANSLATION>}

\promptexample{Prompt for Translation}{
You are a professional translator.
You are given a source text in {SOURCE LANGUAGE}.
You need to translate the source text to {TARGET LANGUAGE}.
Don't include any other text except the translation.
Please output the translation between <START OF TRANSLATION> and <END OF TRANSLATION>. Source text: {SOURCE TEXT}
}

\promptexample{Prompt for LLM-as-a-Testset for Chatbot Arena}{
You are tasked with creating a diverse and engaging prompt for a chatbot arena. This prompt will be used to test and compare the capabilities of different language models. Your goal is to generate a question or prompt that will challenge these models and showcase their strengths or weaknesses.

Here are the input variables you will use to craft your prompt:\\
- Language: language\\
- Topic: topic\\
- Subtopic: subtopic\\
- Difficulty: difficulty\\
- Style: style\\
- Writer: writer\\
- Writing proficiency: writing\_proficiency\\
- Prompt length: length\\

Guidelines for creating the prompt:

1. Abide strictly by the input variables provided.

2. Ensure that your prompt is open-ended enough to allow for varied and interesting responses from different language models.

3. Avoid prompts that are overly specific to a particular AI model's capabilities or training data.

4. Create a prompt that is engaging and thought-provoking, encouraging creative or analytical thinking.

Generate output in the following format:

<START OF PROMPT>
[Your generated prompt here. IMPORTANT: include only the prompt.]
<END OF PROMPT>

Ensure the generated prompt is in the requested language. Remember to abide strictly by the provided input variables and the requested format.
}

\promptexample{An example question for Chatbot Arena (English)}{
In Albion's ledger, where fortunes entwine, what unseen current shapes the City's final line?
}

\promptexample{An example question for Chatbot Arena (Bemba)}{
Finshi fingacitika mu fashion design mu Zambia mu myaka ikumi iileisa? Ndiye, muli busy sana, lelo ndefwaya ukwishiba ifyo mwingalanda pali ifi.
}

\promptexample{Prompt for LLM-as-an-Evaluator  for Chatbot Arena} {You are an expert judge evaluating response quality for prompts on a variety of topics. You will be presented with:

- An original prompt\\
- A response to evaluate\\

Rate the response on a scale of 1-6 based on these key criteria:

- Accuracy and factual correctness\\
- Comprehensiveness of the answer\\
- Clarity and organization\\
- Relevance to the prompt\\
- Supporting evidence/examples (if applicable)\\
- Objectivity and balance (if applicable)\\

Scoring Rubric:\\
6 - Outstanding\\

- Exceptionally accurate and well-researched information\\
- Comprehensive coverage with valuable additional context\\
- Perfectly structured and crystal clear explanation\\
- Directly addresses all aspects of the prompt\\
- Strong supporting evidence and examples\\
- Maintains perfect objectivity where needed\\

5 - Excellent\\

- Highly accurate information\\
- Thorough coverage of the topic\\
- Very clear and well-organized\\
- Closely aligned with the prompt\\
- Good supporting evidence\\
- Strong objectivity\\

4 - Good\\

- Generally accurate information\\
- Adequate coverage of main points\\
- Clear organization with minor issues\\
- Mostly relevant to the prompt\\
- Basic supporting evidence\\
- Reasonable objectivity\\

3 - Fair\\

- Some inaccuracies present\\
- Incomplete coverage\\
- Organization needs improvement\\
- Partially misses prompt intent\\
- Limited supporting evidence\\
- Some bias present\\

2 - Poor\\

- Multiple inaccuracies\\
- Superficial coverage\\
- Unclear organization\\
- Largely misses prompt intent\\
- Minimal supporting evidence\\
- Clear bias present\\

1 - Inadequate\\

- Fundamentally incorrect information\\
- Severely incomplete\\
- Disorganized or incomprehensible\\
- Fails to address prompt\\
- No supporting evidence\\
- Extremely biased or misleading\\

Format your output as follows: \\
Put detailed explanation between <START OF FEEDBACK> and </END OF FEEDBACK>\\
Put result between <START OF RESULT> and </END OF RESULT>\\
Don't provide any other text\\

<START OF FEEDBACK>\\
Put detailed explanation of the score based on the criteria here\\
</END OF FEEDBACK>\\
<START OF RESULT>\\
Put only a number from 1 to 6 here\\
</END OF RESULT>\\

<START OF PROMPT>\\
prompt\\
<END OF PROMPT>\\

<START OF ANSWER>\\
answer\\
<END OF ANSWER>\\}

\begin{table*}[ht]
\centering
\small
\begin{tabular}{l@{\hspace{1mm}}lp{1cm}p{1cm}p{1cm}}
% \toprule
& En${\rightarrow}$Bemba
& \multicolumn{3}{c}{chrF across prompts + models} \\
&& \textbf{Gemini} & \textbf{GPT} & \textbf{Claude} \\
\cmidrule{2-5}
\parbox[t]{2mm}{\multirow{3}{*}{\rotatebox[origin=r]{90}{Data from\hspace{-5mm}}}}
& \bf Gemini & \cellcolor{SeaGreen3!60}36.07 & \cellcolor{SeaGreen3!55}35.77 & \cellcolor{SeaGreen3!30}34.13 \\
& \bf GPT & \cellcolor{SeaGreen3!25}32.98 & \cellcolor{SeaGreen3!60}34.86 & \cellcolor{SeaGreen3!20}31.50 \\
& \bf Claude & \cellcolor{SeaGreen3!60}36.75 & \cellcolor{SeaGreen3!55}36.47 & \cellcolor{SeaGreen3!60}36.85 \\
\bottomrule
\end{tabular}
\quad
\begin{tabular}{l@{\hspace{1mm}}lp{1cm}p{1cm}p{1cm}}
& En${\rightarrow}$Aymara
& \multicolumn{3}{c}{chrF across prompts + models} \\
&& \textbf{Gemini} & \textbf{GPT} & \textbf{Claude} \\
\cmidrule{2-5}
\parbox[t]{2mm}{\multirow{3}{*}{\rotatebox[origin=r]{90}{Data from\hspace{-5mm}}}}
& \bf Gemini & \cellcolor{SeaGreen3!60}36.15 & \cellcolor{SeaGreen3!55}35.89 & \cellcolor{SeaGreen3!30}34.07 \\
& \bf GPT & \cellcolor{SeaGreen3!25}32.83 & \cellcolor{SeaGreen3!60}35.28 & \cellcolor{SeaGreen3!20}31.27 \\
& \bf Claude & \cellcolor{SeaGreen3!60}36.83 & \cellcolor{SeaGreen3!55}36.62 & \cellcolor{SeaGreen3!60}37.14 \\
\bottomrule
\end{tabular}
\small
\caption{Average chrF@K similarity scores, differentiating within-model (diagonal) from cross-model (off-diagonal) comparisons. Unlike XX${\rightarrow}$En translation, a clear diagonal trend is not observed. For instance, Claude-generated outputs exhibit cross-model chrF similarities with GPT4.1 and Gemini-2.5-pro that are comparable to its within model similarity.}
\label{tab:biased_chrf_table_en_xx}
\end{table*}

% \section{Why does Translation Asymmetry Exist for Self-Bias?}
% \label{sec:trans_asy_human_ttr}
% In \Cref{tab:tab_ttr_en_xx_xx_en_comparisons_human_vs_models}, we showed that model-generated English source texts have TTR distributions more similar to human-written English than model-generated texts in other languages (for XX${\rightarrow}$En translation) do to their human-written counterparts. This indicates that the lexical diversity of model-generated English source text is closer to that of human-written English source text.
%
% \begin{table*}[ht]
% \centering
% \small
% \begin{tabular}{lrrr}
% \toprule
%  & \multicolumn{3}{c}{TTR distribution difference: model vs.\ human (Src Only)} \\
% \cmidrule(r){2-4}
% Lang dir & \textbf{Gem\&Hu} & \textbf{GPT\&Hu} & \textbf{Cla\&Hu} \\
% \midrule
% En${\rightarrow}$XX & 1.795 & 1.688 & 1.590 \\
% XX${\rightarrow}$En & \textbf{3.187} & \textbf{2.335} & \textbf{2.232} \\
% \bottomrule
% \end{tabular}
% \caption{TTR distribution difference between model-generated and human-written source texts (source-only generation). Model-generated English source texts (En${\rightarrow}$XX) are much closer to human-written English in lexical diversity than model-generated low-resource texts (XX${\rightarrow}$En) are to their human counterparts, confirming that English generation leaves minimal model-specific signatures.}
% \label{tab:tab_ttr_en_xx_xx_en_comparisons_human_vs_models}
% \end{table*}

\section{Self-repair Degeneration in Translation}
\label{sec:self_repair_degenerated}
In \Cref{tab:self_repair_degenerate}, we examine Gemini-2.5-pro's self-repair ability when translating its own degenerated source texts. Gemini-2.5-pro more effectively corrects degenerated content it produces during the translation process compared to other models, across both language directions. The observed non-deterministic behavior of GPT-4.1 is likely due to the significantly smaller sample of degenerated source sentences it processed (n=23/400) compared to Gemini-2.5-pro (n=89/400). An example of this self-repair during translation is provided below.

\begin{table*}[ht]
\centering
% Vilém: try not to use resizebox to preserve font size consistency
% \resizebox{\textwidth}{!}{
\small
\begin{tabular}{lrrrr}
\toprule
 & \multicolumn{4}{c}{Degeneration ratio after translating on degenerated source texts} \\
 & \multicolumn{2}{c}{\textbf{Aymara$\rightarrow$English}} & \multicolumn{2}{c}{\textbf{Bemba$\rightarrow$English}} \\
\cmidrule(r){2-3} \cmidrule(l){4-5}
Translator & Gemini-2.5-Pro & GPT4.1 & Gemini-2.5-Pro & GPT4.1 \\
\midrule
Gemini-2.5-Pro & \textbf{90.4} & 86.7 & \textbf{81.1} & 75.0\\
GPT4.1 & 94.2 & 93.3 & 86.5 & \textbf{62.5}\\
Claude-Opus-4 & 94.2 & \textbf{73.3} & 94.6 & 87.5\\
\bottomrule
\end{tabular}
% }
\caption{Model Self-Repair Ability during Translation of Self-Generated Degenerated Source Texts.
This table presents the percentage of translations that retain degenerated content. Gemini-2.5-pro demonstrates a consistently superior ability to correct degenerated content in its self-generated source texts during the translation process, outperforming other models across both language directions. GPT-4.1's observed non-deterministic behavior is potentially attributable to the significantly smaller sample size of degenerated source sentences it processed (n=23/400), compared to Gemini-2.5-pro (n=89/400).} 
\label{tab:self_repair_degenerate}
\end{table*}

\promptexample{Example of Gemini's self repair during translation}{\\\\
\textbf{Gemini's degenerate source text:}
Jichhürunakanxa, celularanakax wali wakiskiripuniwa. Aka tecnología ukax janiw mayni jaqimpi aruskipt’añatakikiti, jan ukasti yatiñanak jikxatañataki, anatañataki, ukat yaqha lurawinak lurawinak lurawinak lurawinak lurawinak lurawinak lurawinak lurawinak lurawinak lurawinak lurawinak lurawinak lurawinak lurawinak lurawinak lurawinak lurawinak lurawinak lurawinak lurawinak lurawinak lurawinak lurawinak lurawinak lurawinak lurawinak [repeated 50 times] lurawinak lurawinak lurawinak lurawinak lurawinak lurawinak lurawinak lurawinak lurawinak lurawinak lurawinak lurawinak lurawinak lurawinak lurawinak lurawinak lurawinak lurawinak lurawinak lurawinak lurawinak lurawinak lurawinak lurawinak lurawinak lurawinak lurawinak lurawinak lurawinak lurawinak lurawinak lurawinak lurawinak lurawinak lurawinak lurawinak lurawinak lurawinak lurawinak lurawinak lurawinak lurawinak lurawinak lurawinak lurawinak lurawinak lurawinak lurawinak lurawinak lurawinak lurawinak lurawinak lurawinak lurawinak lurawinak lurawinak lurawinak lurawinak luraw\\

\textbf{Gemini's translation:} Nowadays, cell phones are truly necessary. This technology is not only for communicating with other people, but also for finding information, for entertainment, and for carrying out various other activities.
}

% NOTE: The "Extension to Non-translation Tasks" section has been moved to the main paper (Section \ref{sec:non_translation}).

\begin{table}[ht]
\centering
\small
\begin{tabular}{lccc}
\toprule
\textbf{Translator} \textbackslash\ \textbf{Benchmark} & \bf Gemini & \bf Claude & \bf GPT-4.1 \\
\midrule
Gemini      & \textbf{6.000} & 5.584 & 5.980 \\
Claude      & 5.998 & \textbf{5.633} & 5.986 \\
GPT-4.1     & 5.998 & 5.542 & \textbf{5.987} \\
\bottomrule
\end{tabular}
\caption{LLM-as-a-benchmark scores for Chinese$\to$English translation (1--6 scale; higher is better). All scores are near-perfect, leaving minimal room for self-bias to manifest. The diagonal entries (bolded) are highest in each column, indicating residual self-preference, but the absolute differences are negligible compared to low-resource settings.}
\label{tab:zh_en_scores}
\end{table}

\section{High-Resource Language Analysis}
\label{sec:high_resource}

We considered high-resource languages for the analysis as well. However, the focus of our study is on medium- and low-resource machine translation, primarily due to the near-perfect performance of frontier models in high-resource settings. As shown in \Cref{tab:zh_en_scores} for Chinese-to-English translation, the scores from Gemini, Claude, and GPT-4.1 for all translations are nearly perfect (on a 1--6 scale). While models still display self-bias in their own generation---each model assigns itself the highest score in its column---the magnitude is quantitatively much smaller than in low-resource settings.

We argue that the near-perfect translation quality of all models makes studying self-bias on high-resource languages less informative: there is little room for diverse error patterns, and the differences between self-scores and peer-scores are marginal. Consequently, we prioritize medium- and low-resource language directions, where models exhibit distinct error patterns and evaluate them differently due to self-bias. More importantly, the \llmbench setup is most practically relevant for low-resource translation, as these tasks inherently lack the data to construct high-quality benchmarks through human curation.

\section{LLM-as-an-Evaluator Self-Bias}
\label{sec:evaluator_bias_appendix}

For completeness, we report evaluator-only self-bias. In this setting, source texts are drawn from the human-authored FLORES-200 benchmark, so the LLM acts solely as an evaluator. Since the test set differs from the LLM-generated inputs used in \llmtest and \llmbench, this condition is not directly comparable and is reported separately.

\begin{table}[h]
    \centering
    \small
    \begin{tabular}{lm{9mm}m{9mm}m{9mm}}
    \toprule
    XX${\rightarrow}$En & \bf Gemini & \bf GPT & \bf Claude \\
    \midrule
    \bf \llmeval   
    & -0.302 
    & -0.443 
    & -0.303 \\
    \bottomrule
    \end{tabular}
    \caption{Self-bias under \llmeval (averaged XX$\rightarrow$En), where source texts come from human-authored FLORES-200. All models show evaluator self-preference, consistent with prior work~\citep{xu-etal-2024-pride, llm-favors-its-own-outputs}.}
    \label{tab:evaluator_bias_appendix}
\end{table}

\section{Diversity Filtering Under \llmbench}
\label{sec:diversity_benchmark_appendix}

\Cref{tab:diversity_self_bias_benchmark} reports the full \llmbench self-bias for the diversity-filtered subsets. Source texts with high within-model similarity (Max chrF) consistently exhibit the highest self-bias. Min chrF shows significantly lower self-bias than Max chrF for Gemini and Claude ($p<0.05$), but the difference is not significant for GPT, indicating that the effectiveness of diversity filtering at the full benchmark level is model-dependent.

\begin{table}[h]
\centering
\small
\begin{tabular}{l@{\hspace{1mm}}lm{11mm}m{11mm}m{11mm}}
% \toprule % Using booktabs rules (optional, but good practice)
& 
& \multicolumn{3}{c}{Self-bias Estimation} \\
& Subset & \textbf{Gemini} & \textbf{GPT} & \textbf{Claude} \\
\cmidrule{2-5}
& \bf Max chrF & {-}0.400$^*$ & {-}0.280 & {-}0.685$^*$ \\
& \bf Random  & {-}0.342 & \bf{-}0.256 & {-}0.616\\
& \bf Min chrF  & \bf{-}0.250 & {-}0.265 & \bf{-}0.600 \\
\bottomrule % Using booktabs rules
\end{tabular}
\caption{Self-bias of \llmbench for subsets of LLM-generated source texts selected by within-model chrF@K similarity. Min chrF shows significantly lower self-bias than Max chrF for Gemini and Claude ($p<0.05$), but the difference is not significant for GPT. Self-bias is averaged for Aymara$\rightarrow$En and Bemba$\rightarrow$En.}
\label{tab:diversity_self_bias_benchmark}
\end{table}
%%%%%%%%%%%%%%%%%%%%%%%%%%%%%%%%%%%%%%%%%%%%%%%%%%%%%%%%%%%%%%%%%%%%%%%%%%%%%%%
%%%%%%%%%%%%%%%%%%%%%%%%%%%%%%%%%%%%%%%%%%%%%%%%%%%%%%%%%%%%%%%%%%%%%%%%%%%%%%%

\end{document}